# Quantum Annealing for Variational Bayes Inference

**Issei Sato**
Information Science and Technology
University of Tokyo, Japan

**Kenichi Kurihara**
Google
Tokyo, Japan

**Shu Tanaka**
Institute for Solid State Physics,
University of Tokyo, Japan

**Hiroshi Nakagawa**
Information Technology Center,
The University of Tokyo, Japan

**Seiji Miyashita**
Dept. of Physics,
The University of Tokyo, Japan

## Abstract

This paper presents studies on a deterministic annealing algorithm based on quantum annealing for variational Bayes (QAVB) inference, which can be seen as an extension of the simulated annealing for variational Bayes (SAVB) inference. QAVB is as easy as SAVB to implement. Experiments revealed QAVB finds a better local optimum than SAVB in terms of the variational free energy in latent Dirichlet allocation (LDA).

## 1 Introduction

Several studies that are related to machine learning with quantum mechanics have recently been conducted. The main idea behind these has been based on a generalization of the probability distribution obtained by using a density matrix, which is a self-adjoint positive-semidefinite matrix of trace one. Wolf (2006) connects the basic probability rule of quantum mechanics, called the "Born Rule", which formulates a generalized probability by using a density matrix, to spectral clustering and other machine learning algorithms based on spectral theory. Crammer and Globerson (2006) combined a margin maximization scheme with a probabilistic modeling approach by incorporating the concepts of quantum detection and estimation theory (Helstrom, 1969). Tanaka and Horiguchi (2002) proposed a quantum Markov random field using a density matrix and quantum mechanics and applied to image restoration.

Generalizing a Bayesian framework based on a density matrix has also been proposed. Schack et al. (2001) proposed a "quantum Bayes rule" for conditional density between two probability spaces. Warmuth et al. generalized the Bayes rule to treat a case where the prior was a density matrix (Warmuth, 2005) and unified Bayesian probability calculus for density matrices with rules for translation between joints and conditionals (Warmuth, 2006). Typically, the formulas derived by quantum mechanics generalization have retained the conventional theory as a special case when the density matrices have been diagonal. Computing the full posterior distributions over model parameters for probabilistic graphical models, e.g. latent Dirichlet allocation (Blei et al., 2003), remains difficult in these quantum Bayesian frameworks, as well as classical Bayesian frameworks. In this paper, we generalize the variational Bayes inference (Attias, 1999), which is widely used framework for probabilistic graphical models, based on ideas that have been used in quantum mechanics.

Variational Bayes (VB) inference has been widely used as an approximation of Bayesian inference for probabilistic models that have discrete latent variables. For example, in a probabilistic mixture model, such as a mixture of Gaussians, each data point is assigned to a latent class, and a latent variable corresponding to a data point indicates the latent class. VB is an optimization algorithm that minimizes the cost function. The cost function, called the negative variational free energy, is a function of latent variables. We have called the cost function "energy" in this paper.

Since VB is a gradient algorithm similar to the Expectation Maximization (EM) algorithm, it suffers from a local optimal problem in practice. Deterministic annealing (DA) algorithms have been proposed for the EM algorithm (Ueda and Nakano, 1995) and VB (Katahira et al., 2008) based on simulated annealing (SA) (Kirkpatrick et al., 1983) to overcome issue with local optima. We called simulated annealing based VB SAVB. SA is one of the most well known physics based approaches to machine learning. SA is based on the concept of statistical mechanics, called "temperature". We decrease the parameter of "temperature" gradually in SA. Because the energy landscape becomes flat at high temperature, it is easy to change the state (see Fig.1(a)). However, the state is trapped at low temperature because of the valley in the energy barrier and the transition probability becomes very low. Therefore, SA does not necessarily find a global optimum in the practical cooling schedule of temperature $T$. In physics, quantum annealing (QA) has attracted attention as an alternative annealing method of optimization problems by a process that is analogous to quantum fluctuations (Apolloni et al., 1989; Kadowaki and Nishimori, 1998; Santoro et al., 2002). QA is expected to help states avoid being trapped by poor local optima at low temperatures.

The main point of this paper is to explain the novel DA algorithm for VB based on the QA (QAVB) we derived and present the effects of QAVB we obtained through experiments. QAVB is a generalization of VB and SAVB attained by using a density matrix. We describe our motivation for deriving QAVB in terms of a density matrix in Section 3. Here, we overview the QAVB that we derived. Interestingly, although QAVB is generalized and formulated by a density matrix, the algorithm for QAVB we finally derived does not need operations for a density matrix such as eigenvalue decomposition and only has simple changes from the SAVB algorithm.

Since SAVB does not necessarily find a global optimum, we still need to run multiple SAVBs independently with different random initializations where $m$ denote the number of SAVBs. Here, let us consider running dependently, not independently, multiple SAVBs where "dependently" means that we run multiple SAVBs introducing interaction $f$ among neighboring SAVBs that are randomly numbered such as $j-1$, $j$ and $j+1$ (see Fig.1(b)). In Fig.1, $\sigma_j$ indicates the latent class states of $N$ data points in the $j$-th SAVB. The independent SAVBs have a very low transition probability among states, i.e., they have been trapped, at high temperature as shown in Fig.1(c), while the dependent QAVBs can changes the state in that situation. This is because interaction $f$ starts from zero (i.e., "independent"), gradually increases, and makes $\sigma_{j-1}$ and $\sigma_j$ approach each other, the state will then be moved into $\sigma^*$. If there is a better state around sub-optimal states that the independent SAVBs find, the dependent SAVBs are expected to work well. The dependent SAVBs are just QAVB where interaction $f$ and the above scheme are derived from QA mechanisms as will be explained in the following section.

This paper is organized as follows. In Section 2, we introduce the notations used in this paper. In Section 3, we motivate QAVB in terms of a density matrix. Section 4 and 5 explain how we derive QAVB and present the experimental results in latent Dirichlet allocation (LDA). Section 6 concludes this paper.

## 2 Preliminaries

We assume that we have $N$ data points, and they are assigned to $K$ latent classes. The latent class of the $i$-th data point is denoted by the latent variable $z_i$. $z_i = k$ indicates that the latent class of the $i$-th data point is $k$. The latent class of the $i$-th data point is also denoted by $K$ dimensional binary indicator vector $\tilde{\sigma}_i$ where if $z_i$ is equal to $k$, the $k$-th element of $\tilde{\sigma}_i$ is equal to 1 and the other elements are all equal to 0. The number of available class assignment of all data points is $K^N$. The class assignment of all data points is denoted by $K^N$ dimensional binary indicator vector $\sigma = \bigotimes_{i=1}^N \tilde{\sigma}_i$ where $\bigotimes$ is the Kronecker product, which is a special case of a tensor product. If $A$ is $k$-by-$l$ matrix and B is an $m$-by-$n$ matrix, then the Kronecker product $A \bigotimes B$ is the $km$-by-$ln$ block matrix as follows: $A \bigotimes B = \begin{pmatrix} a_{11}B & \cdots & a_{1l}B \\ \vdots & \ddots & \vdots \\ a_{k1}B & \cdots & a_{kl}B \end{pmatrix}$. For example, if $K = 2$, $N = 2$, $z_1 = 1$ ($\tilde{\sigma}_1 = (1,0)^T$) and $z_2 = 2$ ($\tilde{\sigma}_2 = (0,1)^T$), then $\sigma = \tilde{\sigma}_1 \bigotimes \tilde{\sigma}_2 = (0,1,0,0)^T$.

Let $\boldsymbol{x} = (\boldsymbol{x}_1, \cdots, \boldsymbol{x}_N)$ denote the $N$ observed data points and $\boldsymbol{\theta}$ denote the model parameters. $\sigma^{(l)}$ indicates the $l$-th latent class states of $K^N$ available latent class states. For example, if $K = 2$ and $N = 2$, then $\sigma^{(1)} = (1,0,0,0)^T$, $\sigma^{(2)} = (0,1,0,0)^T$, $\sigma^{(3)} = (0,0,1,0)^T$ and $\sigma^{(4)} = (0,0,0,1)^T$. The set of available latent class states is denoted by $\Sigma = \{\sigma^{(l)} | (l = 1, 2, \cdots, K^N)\}$.

## 3 Motivation for QAVB in terms of Density matrix

For those unfamiliar with quantum information processing, we will explain a density matrix which can be used as an extension of conventional probability. Our definition of a density matrix is based on (Warmuth, 2006).

A density matrix is a self-adjoint positive-semidefinite matrix and its trace is one. Conventional probability which we called *classical* statistics can be expressed by a diagonal density matrix as follows. For example, let us consider the case of two data points and two latent classes as well as Section 2. We define four states, denoted by indicator vectors $\{\sigma^{(i)}\}_{i=1}^4$, and probability vector $\mathbf{p} = (p_1, p_2, p_3, p_4)^T$, where $p_i$ indicates the occurrence probability of the $i$-th state $\sigma^{(i)}$.

Then, the density matrix of this system is given by

$$diag\{p_1, p_2, p_3, p_4\} = \sum_{i=1}^4 p_i \sigma^{(i)} \sigma^{(i)T}, \tag{1}$$

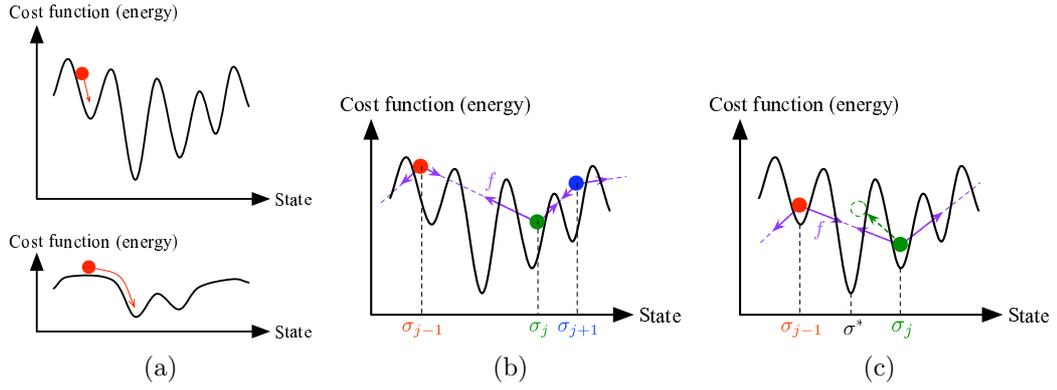

Figure 1: (a) Schematic picture of SAVB. (Upper panel) At low temperature, the state often falls into local optima. (Bottom panel) At high temperature, since the energy landscape becomes flat, the state can change over a wide range. (b) and (c) Schematic picture of QAVB. (b) QAVB connects neighboring SAVBs. (c) $\sigma_j$ can reach $\sigma^*$ owing to the interaction $f$. It seems to go through energy barrier.

where $diag\{\cdot\}$ indicates diagonal matrix. We can extend the concept of probability by introducing non-diagonal elements in a density matrix which is called *quantum* statistics. A state of a system in quantum statistics is defined by a unit (column) real vector[1], $\boldsymbol{u}$, where dyad $\boldsymbol{u}\boldsymbol{u}^T$ has trace one, $\text{Tr}\left(\boldsymbol{u}\boldsymbol{u}^T\right) = \text{Tr}\left(\boldsymbol{u}^T\boldsymbol{u}\right) = 1$. A density matrix, $\Phi$, generalizes a finite probability distribution and can be defined as a mixture of dyads,

$$\Phi = \sum_i p_i \boldsymbol{u}_i \boldsymbol{u}_i^T, \quad (2)$$

where $p_i$ is a mixture proportion (coefficient) that is non-negative and sums to one. $p_i$ specifies the proportion of the system in state $\boldsymbol{u}_i$. A density matrix assigns a probability to the unit vector or its associated dyad given by $p(\boldsymbol{u}) = \text{Tr}\left(\Phi \boldsymbol{u}\boldsymbol{u}^T\right) (= \boldsymbol{u}^T \Phi \boldsymbol{u})$. This is called the "Born rule" in quantum mechanisms. According to Gleason's theorem, there is a one to one correspondence between generalized probability distributions and density matrices (Gleason, 1957). For example, when a state vector is $\boldsymbol{u} = \left(\frac{1}{2}, 0, \frac{\sqrt{3}}{2}, 0\right)$, it represents the mixture of the first state and the third state with probability $\left(\frac{1}{2}\right)^2 = \frac{1}{4}$ and $\left(\frac{\sqrt{3}}{2}\right)^2 = \frac{3}{4}$, respectively.

A probabilistic model employs uncertainty to model phenomena, and has demonstrated its practically in many scientific fields. Although classical statistics involves uncertainty over mixture proportions ($\{p_i\}$), it restricts state vectors to indicator vectors ($\{\sigma^{(i)}\}$). In contrast, quantum statistics involves uncertainty over not only mixture proportions ($\{p_i\}$) but also state vectors ($\{\boldsymbol{u}_i\}$) because if density matrix $\Phi$ has off-diagonal elements, state vectors $\{\boldsymbol{u}_i\}$ take arbitrary vectors. Therefore, a probabilistic model based on quantum statistics is a more generalized model in terms of uncertainty, and the generalization is expected to be more useful. In the same way, since classical VB inference including SA variants only involves uncertainty over mixture proportions, this paper proposes a method of maintaining uncertainty over state vectors.

Finally, Fig 2 sums up the relationship between VB, SAVB, and QAVB in terms of a density matrix. SAVB and QAVB control uncertainty of mixture proportions via temperature $T$. However, QAVB can control the uncertainty of state vectors by introducing quantum effect parameter $\Gamma$ that is described in Section 4, leading to enhanced generalization.

## 4 Quantum Annealing for Variational Bayes Inference

This section explains how we derive update equations for QAVB. First, we define the lower bound of the marginal likelihood in QAVB as typical VB. Then, we apply Suzuki-Trotter expansion (Trotter, 1959; Suzuki, 1976) to the marginal of QA to analytically obtain update equations.

### 4.1 Introducing Quantum Effect

We define $\mathcal{H}_c$ with a $K^N$ by $K^N$ diagonal matrix as follows:

$$\mathcal{H}_c = diag\{-\log p(\boldsymbol{x}, \sigma^{(1)}), \cdots, -\log p(\boldsymbol{x}, \sigma^{(K^N)})\} \quad (3)$$

---
[1] A state vector generally does not need to be a restricted real vector. If we consider a complex vector, the definition of the trace of a dyad is replaced by $\text{Tr}(\boldsymbol{u}\boldsymbol{u}^*) = \text{Tr}(\boldsymbol{u}^*\boldsymbol{u}) = 1$, where $\boldsymbol{u}^*$ indicates complex conjugate of $\boldsymbol{u}$. However, for simplicity, we have restricted the real vector in this paper.

| Method \ Uncertainty | Mixture proportions $p_i$ (Conventional fields) | State vectors $\mathbf{u}_i$ (Target field in this paper) |
|---|---|---|
| VB (Attias, 1999) | ✓ | ✗ |
| SAVB (Katahira, 2008) | ✓ | ✗ |
| **QAVB** (This study) | ✓ | ✓ |

Figure 2: The uncertainty over mixture proportions has been well studied in machine learning. VB and SAVB also only involve uncertainty over mixture proportions. We study the uncertainty over another component of a density matrix, state vectors. QAVB involves uncertainty over not only mixture proportions but also state vectors.

The conditional probability of indicator state vector $\sigma$ given $\boldsymbol{x}$ is calculated by

$$p(\sigma|\boldsymbol{x}) = \frac{p(\boldsymbol{x},\sigma)}{p(\boldsymbol{x})} = \frac{\sigma^T e^{-\mathcal{H}_c}\sigma}{\text{Tr}\,(e^{-\mathcal{H}_c})} = \sigma^T \Phi_c \sigma = \text{Tr}\,(\Phi_c \sigma \sigma^T), \quad (4)$$

where $\Phi_c = \frac{e^{-\mathcal{H}_c}}{\text{Tr}\,(e^{-\mathcal{H}_c})}$ is a density matrix.

The marginal log-likelihood of $N$ data points is formulated as

$$\log p(\boldsymbol{x}) = \log \text{Tr}\{e^{-\mathcal{H}_c}\}. \quad (5)$$

Since the fully conditional posteriors are intractable, VB inference is proposed as an approximated algorithm for estimating conditional posteriors.

The marginal log-likelihood of $p(\boldsymbol{x})$ can be lower bounded by introducing distribution over latent variables $\sigma$, parameters $\boldsymbol{\theta}$ and the approximate distribution $q(\sigma)q(\boldsymbol{\theta})$ of a posteriori distribution $p(\sigma,\boldsymbol{\theta}|\boldsymbol{x})$ as follows.

$$\log p(\boldsymbol{x}) \geq \sum_\sigma \int q(\sigma)q(\boldsymbol{\theta}) \log \frac{p(\boldsymbol{x},\sigma,\boldsymbol{\theta})}{q(\sigma)q(\boldsymbol{\theta})} d\boldsymbol{\theta} \quad (6)$$

$$= \tilde{F}[q(\sigma),q(\boldsymbol{\theta})]. \quad (7)$$

We maximize $\tilde{F}[q(\sigma),q(\boldsymbol{\theta})]$ with respect to $q(\sigma)q(\boldsymbol{\theta})$ to obtain a better approximation of $p(\sigma,\boldsymbol{\theta}|\boldsymbol{x})$ in VB inference. $\tilde{F}[q(\sigma),q(\boldsymbol{\theta})]$ is called the variational free energy.

We derive QAVB by maximizing the lower bound of the following marginal log-likelihood.

$$\log p(\boldsymbol{x};\beta,\Gamma) = \log \text{Tr}\{e^{-\beta\mathcal{H}}\}, \quad (8)$$

where $\Gamma$ is the quantum effect parameter, $\beta$ is inverse temperature, i.e., $\beta = \frac{1}{T}$, and we define $\mathcal{H}$ with a $K^N$ by $K^N$ matrix as follows:

$$\mathcal{H} = \mathcal{H}_c + \mathcal{H}_q, \quad (9)$$

$$\mathcal{H}_q = \sum_{i=1}^N \sigma_{xi}, \quad \sigma_{xi} = \left(\bigotimes_{j=1}^{i-1} \mathbb{E}_K\right) \otimes \sigma_x \otimes \left(\bigotimes_{l=i+1}^N \mathbb{E}_K\right),$$

$$\sigma_x = \Gamma(\mathbb{E}_K - 1_K), \quad (10)$$

where $\mathbb{E}_K$ is the $K$ by $K$ identity matrix, $1_K$ is the $K$ by $K$ matrix whose elements are all one, and $\mathcal{H}_q$ is a symmetrical matrix. The above $\mathcal{H}$ is a standard setting for QA (Kadowaki and Nishimori, 1998). The conditional probability of $\sigma$ given $\boldsymbol{x}$, $\beta$ and $\Gamma$ is calculated by

$$p(\sigma|\boldsymbol{x};\beta,\Gamma) = \frac{\sigma^T e^{-\beta\mathcal{H}}\sigma}{\text{Tr}\,(e^{-\beta\mathcal{H}})} = \sigma^T \Phi_q \sigma = \text{Tr}\,(\Phi_q \sigma \sigma^T), \quad (11)$$

where $\Phi_q = \frac{e^{-\beta\mathcal{H}}}{\text{Tr}\,(e^{-\beta\mathcal{H}})}$ is a density matrix.

Note that $\mathcal{H}$ becomes diagonal if $\Gamma$ is zero, in which case it reduces to $\mathcal{H}_c$, and quantum log-likelihood $\log p(\boldsymbol{x};\Gamma,\beta)$ in Eq. (8) becomes classical loglikelihood $\log p(\boldsymbol{x})$ in Eq. (5) if $\beta$ is one.

The following section explains how we derived an approximated posteriori distributions that maximized the lower bound of $\log p(\boldsymbol{x};\Gamma,\beta)$.

### 4.2 Derivation

Let $\sigma_j$ be one of all the available class assignment states of $N$ data points, s.t. $\sigma_j \in \Sigma$. The class of the $i$-th data point in $\sigma_j$ is denoted by $\tilde{\sigma}_{j,i}$, s.t. $\sigma_j = \bigotimes_{i=1}^N \tilde{\sigma}_{j,i}$. It is intractable to evaluate $\log \text{Tr}\{e^{-\beta\mathcal{H}}\}$ because $\mathcal{H}$ is not diagonal. However, we can approximately trace $e^{-\beta\mathcal{H}}$ by Suzuki-Trotter expansion as follows (see Appendix A) (Suzuki, 1976).

$$p(\boldsymbol{x};\Gamma,\beta) \approx p(\boldsymbol{x};\Gamma,\beta,m) + \mathcal{O}\left(\frac{\beta^2}{m}\right), \quad (12)$$

$$p(\boldsymbol{x};\Gamma,\beta,m) =$$

$$\sum_{\sigma_1} \cdots \sum_{\sigma_m} \prod_{j=1}^m e^{\frac{\beta}{m}\log p(\boldsymbol{x},\sigma_j)} b^N e^{s(\sigma_j,\sigma_{j+1})f(\beta,\Gamma)}, \quad (13)$$

$$s(\sigma_j,\sigma_{j+1}) = \sum_{i=1}^N \delta(\tilde{\sigma}_{j,i},\tilde{\sigma}_{j+1,i}), \quad f(\beta,\Gamma) = \log(\frac{a+b}{b}), \quad (14)$$

$$a = \exp(-\frac{\beta\Gamma}{m}), \quad b = \frac{1}{K}a(a^{-K} - 1), \quad (15)$$

where $\delta(\tilde{\sigma}_{j,i}, \tilde{\sigma}_{j+1,i}) = 1$ if $\tilde{\sigma}_{j,i} = \tilde{\sigma}_{j+1,i}$, and $\delta(\tilde{\sigma}_{j,i}, \tilde{\sigma}_{j+1,i}) = 0$ otherwise. We assume a periodic boundary condition, i.e., $\tilde{\sigma}_{m+1,i} = \tilde{\sigma}_{1,i}$. $m$ is called Trotter number where the above trace can be accurately evaluated within the limit of $m \to \infty$. $\frac{1}{N} s(\sigma_j, \sigma_{j+1})$ indicates a similarity measure that takes [0,1] where $\frac{1}{N} s(\sigma_j, \sigma_{j+1}) = 1$ when $\sigma_j = \sigma_{j+1}$ and $\frac{1}{N} s(\sigma_j, \sigma_{j+1}) = 0$ when $\sigma_j$ and $\sigma_{j+1}$ are completely different.

In the following, we derive the lower bound of $\log p(\boldsymbol{x}; \Gamma, \beta, m)$ by introducing the approximated distributions $q(\sigma_j)$ and $q(\boldsymbol{\theta}_j)$ ($j = 1, \cdots, m$).

$$\log p(\boldsymbol{x}; \Gamma, \beta, m) \geq F_{\text{c}}[m, \beta] + F_{\text{q}}[m, \beta], \quad (16)$$

$$F_{\text{c}}[m, \beta] = \sum_{j=1}^{m} \{ \sum_{\sigma_j} \int q(\sigma_j) q(\boldsymbol{\theta}_j) \left( \log \frac{p(\boldsymbol{x}, \sigma_j, \boldsymbol{\theta}_j)^{\beta_{\text{eff}}}}{q(\sigma_j) q(\boldsymbol{\theta}_j)} \right) d\boldsymbol{\theta}_j \}, \quad (17)$$

$$F_{\text{q}}[m, \beta] = \sum_{j=1}^{m} \sum_{\sigma_j} \sum_{\sigma_{j+1}} q(\sigma_j) q(\sigma_{j+1}) (N \log b + s(\sigma_j, \sigma_{j+1}) f(\beta, \Gamma)), \quad (18)$$

where $\beta_{\text{eff}} = \frac{\beta}{m}$ is called the effective inverse temperature. If $\beta_{\text{eff}} = 1$, $F_{\text{c}}[m, \beta]$ is the sum of $m$ classical variational free energy, i.e., $F_{\text{c}}[m, \beta = 1] = \sum_{j=1}^{m} \tilde{F}[q(\sigma_j), q(\boldsymbol{\theta}_j)]$. $F_{\text{q}}[m, \beta]$ becomes large as $\sigma_j$ and $\sigma_{j+1}$ move approach each other. In practice, the Trotter number $m$ indicates the number of multiple SAVBs with different initializations. $q(\sigma_j)$ and $q(\boldsymbol{\theta}_j)$ are the approximations of posterior distributions in the $j$-th SAVB where index $j = 1, \cdots, m$ is randomly labeled. $f(\beta, \Gamma)$ indicates the interaction between the $j$-th and the $j+1$-th SAVB.

One problem crops up here. The class labels are not always consistent between the $j$-th and the $j + 1$-th SAVB, i.e., class label $k$ in the $j$-th SAVB does not always correspond to class label $k$ in the $j + 1$-th SAVB because the initialization of SAVBs is not the same. For example, assume that $(z_{j,1}, z_{j,2}, z_{j,3}) = (1, 1, 2)$ and $(z_{j+1,1}, z_{j+1,2}, z_{j+1,3}) = (2, 2, 1)$ where $z_{j,i}$ denotes the latent class label of the $i$-th data point in the $j$-th SAVB. In this situation, it can be said that class label 1 in the $j$-th SAVB does not correspond to class label 1 but class label 2 in the $j + 1$-th SAVB.

Let us introduce the projection $\rho_j$ in class labels to absorb the difference of class labels between the $j$-th and the $j+1$-th SAVB. $k' = \rho_j(k)$ indicates that $k$ in the $j$-th SAVB corresponds to $k'$ in the $j+1$-th SAVB. In this way, we have $\delta(\tilde{\sigma}_{j,i}, \tilde{\sigma}_{j+1,i}) = \sum_{k=1}^{K} \sigma_{j,i,k} \sigma_{j+1,i,\rho_j(k)}$ where $\tilde{\sigma}_{j,i} = (\sigma_{j,i,1}, \cdots, \sigma_{j,i,K})$, i.e., $\sigma_{j,i,k}$ takes 1 if $z_{j,i} = k$, and otherwise 0. $q(\sigma_{j,i,k})$ denotes $q(z_{j,i} = k)$.

**Algorithm 1** Quantum Annealing for Variational Bayes Inference.
1: Initialize inverse temperature $\beta_{\text{eff}}$, quantum field $\Gamma$ and model parameters.
2: **for all** iteration $t$ such that $1 \leq t \leq L^{out}$ where $L^{out}$ denotes the number of outer iterations **do**
3:    **for** $j = 1, ..., m$ **do**
4:      **for all** iteration $l$ such that $1 \leq l \leq L^{in}$ where $L^{in}$ denotes the number of inner iterations **do**
5:        **for** $i = 1, ..., N$ **do**
6:          VB-E step: Update $q(\sigma_{j,i})$ with Eq. (20)
7:        **end for**
8:        VB-M step: Update $q(\boldsymbol{\theta}_j)$ with Eq. (21)
9:      **end for**
10:    **end for**
11:    Compute $\rho$ with Eq. (22) and Eq. (23)
12:    Increase inverse temperature $\beta_{\text{eff}}$ (if $\beta_{\text{eff}} > 1$, $\beta_{\text{eff}} = 1$), and decrease quantum field $\Gamma$.
13: **end for**

We have

$$F_{\text{q}}[m, \beta] = mN \log b + f(\beta, \Gamma) \sum_{j=1}^{m} \sum_{i=1}^{N} \sum_{k=1}^{K} q(\sigma_{j,i,k}) q(\sigma_{j+1,i,\rho(k)}). \quad (19)$$

Therefore, we obtain the following updates by taking the functional derivatives of $F_{\text{c}}[m, \beta] + F_{\text{q}}[m, \beta]$ with respect to $q(\sigma_{j,i,k})$ and $q(\boldsymbol{\theta}_j)$, and equating them to zero

$$q(\sigma_{j,i,k}) \propto \exp\{ \int q(\boldsymbol{\theta}_j) \beta_{\text{eff}} \log p(\boldsymbol{x}, \sigma_j, \boldsymbol{\theta}_j) d\boldsymbol{\theta}_j + f(\beta, \Gamma) (q(\sigma_{j-1, i, \rho_{j-1}^{-1}(k)}) + q(\sigma_{j+1, i, \rho_j(k)})) \} \quad (20)$$

$$q(\boldsymbol{\theta}_j) \propto p(\boldsymbol{\theta}_j)^{\beta_{\text{eff}}} \exp\{ \sum_{\sigma_j} q(\sigma_j) \beta_{\text{eff}} \log p(\boldsymbol{x}, \sigma_j, \boldsymbol{\theta}_j) \}, \quad (21)$$

where $\rho^{-1}$ is the inverse projection of $\rho$. $q(\sigma_{j,i,k})$ indicates the probability that the latent class of the $i$-th data point will be $k$ in the $j$-th SAVB. As clarified by Eq. (20), $q(\sigma_{j,i,k})$ approaches $q(\sigma_{j-1,i,\rho_{j-1}^{-1}(k)})$ and $q(\sigma_{j+1,i,\rho_j(k)})$ as $f(\beta, \Gamma)$ Increases. Therefore, $f(\beta, \Gamma)$ works as the interaction explained by Fig 1(b).

### 4.3 Estimates of Class-Label Projection $\rho$

We estimate the class label projection, $\rho$, because such projections represent implicit information. We estimate $\rho$ by maximizing $F_{\text{c}}[m, \beta] + F_{\text{q}}[m, \beta]$ To be more precise, we extract the pairs $(k, \rho_j(k))(j = 1, \cdots, m)$ that maximize $\sum_{j=1}^{m}\sum_{i=1}^{N}\sum_{k=1}^{K} q(\sigma_{j,i,k})q(\sigma_{j+1,i,\rho_j(k)})$ in Eq. (19). This is called the "assignment problem", which is one of the fundamental combinatorial optimization problems. Even though the Hungarian algorithm solves the assignment problem with computational complexity $O(K^3)$, we use the following approximation algorithm whose computational complexity is $O(K^2)$

$$\rho_j(k) = \operatorname*{argmax}_{k'} \sum_{i=1}^{N} q(\sigma_{j,i,k})q(\sigma_{j+1,i,k'}), \quad (22)$$

$$\rho_{j-1}^{-1}(k) = \operatorname*{argmax}_{k'} \sum_{i=1}^{N} q(\sigma_{j,i,k})q(\sigma_{j-1,i,k'}). \quad (23)$$

The $\rho_j$ above means that $k$ in the $j$-th SAVB corresponds to $k'$ in the $j+1$-th SAVB that has the highest correlation between $(q(\sigma_{j,1,k}),\cdots,q(\sigma_{j,N,k}))$ and $(q(\sigma_{j+1,1,k'}),\cdots,q(\sigma_{j+1,N,k'}))$.

## 5 Experiments

We applied SAVB and QAVB to latent Dirichlet allocation (LDA) that is one of the most famous probabilistic graphical models (Blei et al., 2003). We used the Reuters corpus[2] and the Medline corpus[3]. We randomly chose 1,000 documents from the Reuters corpus that had a vocabulary of 12,788 items. We randomly chose 1,000 documents from the Medline corpus that had a vocabulary of 14,252 items. We set the number of topics of LDA to 20.

### 5.1 Annealing schedule

The annealing schedule of temperature $T$ (in practice, inverse temperature $\beta = \frac{1}{T}$) and quantum effect parameter $\Gamma$ exert a substantial influence of SAVB and QAVB processes. Although a certified schedule for temperature is well known in Monte Carlo simulations (Geman and Geman, 1984), we have not yet obtained any mathematically rigorous arguments for $T$ and $\Gamma$ in SAVB and QAVB. Since interaction $f$ is a function of $\Gamma$ and $\beta$, we have to consider the schedule of $f$ in practice.

In this paper, we use the annealing schedule $\beta = \beta_0 r_\beta^t$ and $\beta_{\text{eff}} = \beta_{\text{eff}0} r_{\beta_{\text{eff}}}^t$ that Katahira et al. (2008) used. $t$ denotes the $t$-th iteration.

We also use the following annealing schedule $\Gamma = \Gamma_0 \frac{1}{\sqrt{t}}$ Kadowaki and Nishimori (1998) used. We tried the schedules of $\beta$ with combinations of $\beta_0$=0.2, 0.4, 0.6 and 0.8, and $r_\beta$=1.05, 1.1 and 1.2 in SAVB. As a results, we observed $\beta_0 = 0.6$ and $r_\beta = 1.05$ created an effective schedule in SAVB for LDA. The too low inverse temperature did not work well in LDA. This observation was similar to SAVB for the hidden Markov model (Katahira et al., 2008). Therefore, we set $\beta_0 = \beta_{\text{eff}0} = 0.6$ and $r_\beta = r_{\beta_{\text{eff}}} = 1.05$ in SAVB and QAVB. We varied $\Gamma_0$ and have shown the schedule of $\beta$ and $f$ in Fig.3.

[2] http://www.daviddlewis.com/resources/testcollections/reuters21578/
[3] http://www.nlm.nih.gov/pubs/factsheets/medline.html

### 5.2 Experimental results

We ran QAVB five times in all experiments with a Trotter number, $m$, of 10. The results from this experiment were the average of the minimum negative variational free energy, $\min_j\{-\tilde{F}[q(\sigma_j), q(\boldsymbol{\theta}_j)]\}$, of each run. SAVB was randomly restarted until it consumed the same amount of time as QAVB. We ran five batches of SAVB, and each batch consisted of 20 repetitions of SAVB. The results from this experiment were the average of the minimum variational free energy of all batches. These experimental conditions for QAVB and SAVB enabled a fair comparison of these two experiments in terms of the execution time. In fact, the averaged execution times for QAVB ($m = 10$) and 20 SAVBs corresponds to 20.5 and 22.3 h for Reuters, and 20.4 and 22.9 h for Medline. We set the number of outer iterations at $L^{out} = 300$ in Step 1 in Algorithm 1. The number of inner iterations we tried was $L^{in}$=1, 5, 10 and 20 in SAVB. We found $L^{in} = 20$ was effective in SAVB for LDA. Therefore, we set $L^{in} = 20$ in SAVB and QAVB for LDA.

Fig.4 plots the averages for the minimum negative variational free energy with the mean squared error for Reuters and Medline. In both corpora, each of which has different properties, QAVB outperforms SAVB for each $\Gamma_0$ because the introduction of a novel uncertainty into a model, in this case LDA, works well. QAVB approaches SAVB as $\Gamma_0$ increases because interaction $f$ remains 0 in the limited number of iterations. Moreover, we observed QAVB worked well if interaction $f > 0$ after SAVBs find sub-optimal states. We think fast schedules, i.e. small $\Gamma_0$, did not perform well because the term with interaction $f$ in Eq. (20) is noisy when $q(\sigma)$ is not estimated accurately in the small number of iterations.

## 6 Conclusion

We proposed quantum annealing for variational Bayes inference (QAVB). QAVB is a generalization of the conventional variational Bayes (VB) inference and simulated annealing based VB (SAVB) inference obtained by using a density matrix that generalizes a finite probability distribution. QAVB is as easy as SAVB to implement because QAVB only has to add interaction $f$ to multiple SAVBs, and only one parameter, $\Gamma_0$, is added in practice. The computational complexity of QAVB is larger than that of SAVB because QAVB looks like $m$ parallel SAVBs with interactions. However, we empirically demonstrated that QAVB works better than SAVB which is randomly restarted until it uses the same amount of time as QAVB in latent Dirichlet allocation (LDA). Actually, it is typical to run SAVB many times because SAVB does not necessarily find a global optimum and is trapped by poor local optima at low temperature. In practice, the bottleneck in QAVB is the computational complexity of the projection of class labels in Section 4.3, which is a search problem for one nearest neighbor. An improvement in this algorithm to project class labels would lead to more effective QAVB.

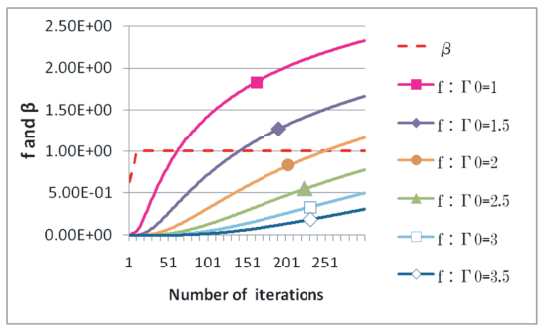

Figure 3: Schedules for inverse temperature $\beta$ and interaction $f$.

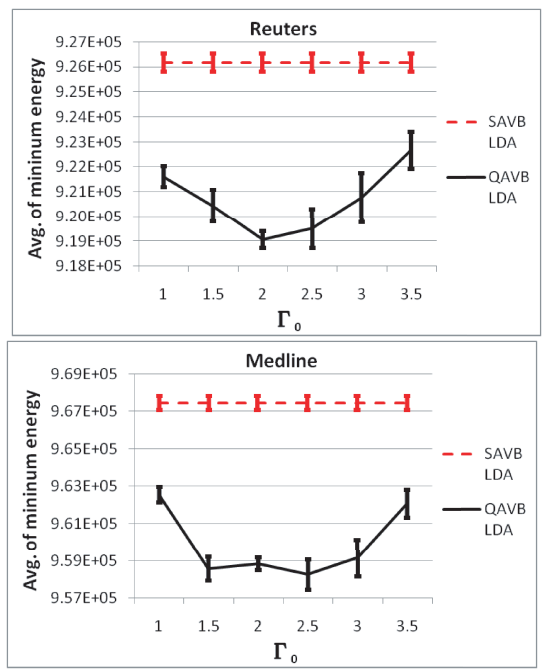

Figure 4: Comparison of QAVB and SAVB in Reuters (Top) and Medline (Bottom). The horizontal axis is $\Gamma_0$. The vertical axis is the average for the minimum energy where the low energy is preferable.

Finally, let us describe future work. We intend to investigate an effective projection algorithm, other constructions of quantum effect $\mathcal{H}_q$, and a suitable schedule of a quantum field for $\Gamma$. We also plan to apply QAVB to other probabilistic models, e.g., a mixture of Gaussians and the hidden Markov model.

## Acknowledgements

This research was funded in part by a MEXT Grant-in-Aid for Scientific Research on Priority Areas "i-explosion" in Japan. This work was partially supported by Research on Priority Areas "Physics of new quantum phases in superclean materials" (Grant No. 17071011) from MEXT, and also by the Next Generation Super Computer Project, Nanoscience Program from MEXT. The authors also thank the Supercomputer Center, Institute for Solid State Physics, University of Tokyo for the use of the facilities.

## A Details of the Suzuki-Trotter Expansion

This section provides the details to derive Eq. (12) from Eq. (10). If $A_1, \cdots, A_n$ are symmetric matrices, the Trotter product formula (Trotter, 1959) is $\exp\left(\sum_{i=1}^{n} A_i\right) = \left(\prod_{i=1}^{n} \exp(A_i/m)\right)^m + \mathcal{O}\left(\frac{1}{m}\right)$. Note that $\left(\prod_{i=1}^{n} \exp(A_i/m)\right)^m$ becomes equal to $\exp\left(\sum_{i=1}^{n} A_i\right)$ in the limit of $m \to \infty$.

Hence, let $\sigma_1$ be the $K^N$-dimensional binary indicator vector mentioned in Section 2. we have

$$\text{Tr}\{e^{-\beta(\mathcal{H}_c + \mathcal{H}_q)}\} = \sum_{\sigma_1} \sigma_1^T e^{-\beta(\mathcal{H}_c + \mathcal{H}_q)} \sigma_1 \quad (24)$$

$$= \sum_{\sigma_1} \sigma_1^T \left(e^{-\frac{\beta}{m}\mathcal{H}_c} e^{-\frac{\beta}{m}\mathcal{H}_q}\right)^m \sigma_1 + \mathcal{O}\left(\frac{\beta^2}{m}\right). \quad (25)$$

Then, by inserting the identity matrices: $\sum_{\sigma_j} \sigma_j \sigma_j^T = \mathbb{E}_{K^N}$ between the product of $m$ exponentials in Eq. (25), $\text{Tr}\{e^{-\beta\mathcal{H}}\}$ leads to

$$\text{Tr}\{e^{-\beta\mathcal{H}}\} = \sum_{\sigma_1} \sum_{\sigma_1'} \cdots \sum_{\sigma_m} \sum_{\sigma_m'} \sigma_1^T e^{-\frac{\beta}{m}\mathcal{H}_c} \sigma_1' \sigma_1'^T e^{-\frac{\beta}{m}\mathcal{H}_q} \sigma_2$$
$$\cdots \sigma_m^T e^{-\frac{\beta}{m}\mathcal{H}_c} \sigma_m' \sigma_m'^T e^{-\frac{\beta}{m}\mathcal{H}_q} \sigma_1. \quad (26)$$

The expression above means auxiliary variables are marginalized out: $\{\sigma_1, \sigma_1', \sigma_2, \sigma_2', ..., \sigma_m, \sigma_m'\}$.

Here, we derive simpler expressions for $\sigma_j^T e^{-\frac{\beta}{m}\mathcal{H}_c} \sigma_j'$ and $\sigma_j'^T e^{-\frac{\beta}{m}\mathcal{H}_q} \sigma_{j+1}$. The former derives the following expression directly from its definition,

$$\sigma_j^T e^{-\frac{\beta}{m}\mathcal{H}_c} \sigma_j' = e^{\frac{\beta}{m} \log p(\boldsymbol{x}, \sigma_j)} \delta(\sigma_j, \sigma_j'), \quad (27)$$

where $\delta(\sigma_j, \sigma_j') = 1$ if $\sigma_j = \sigma_j'$ and $\delta(\sigma_j, \sigma_j') = 0$ otherwise. Next, we derive simpler expression for $\sigma_j'^T e^{-\frac{\beta}{m}\mathcal{H}_q} \sigma_{j+1}$. Using $(A \otimes B)(C \otimes D) = (AC) \otimes (BD)$, $e^{A_1 + A_2} = e^{A_1} e^{A_2}$ when $A_1 A_2 = A_2 A_1$, and $\sigma_j = \bigotimes_{i=1}^{n} \tilde{\sigma}_{j,i}$, we find,

$$\sigma_j'^T e^{-\frac{\beta}{m}\mathcal{H}_q} \sigma_{j+1} = \sigma_j'^T \left(\bigotimes_{i=1}^{n} e^{-\frac{\beta}{m}\sigma_{xi}}\right) \sigma_{j+1}$$
$$= \prod_{i=1}^{N} \tilde{\sigma}'_{j,i}^T e^{-\frac{\beta}{m}\sigma_x} \tilde{\sigma}_{j+1,i},$$
$$= \prod_{i=1}^{N} \tilde{\sigma}'_{j,i}^T \sum_{l=0}^{\infty} \frac{1}{l!} \left(-\frac{\beta}{m}\sigma_x\right)^l \tilde{\sigma}_{j+1,i}$$
$$= \prod_{i=1}^{n} \sum_{l=0}^{\infty} \frac{1}{l!} \left(-\frac{\beta}{m}\right)^l \tilde{\sigma}'_{j,i}^T \sigma_x^l \tilde{\sigma}_{j+1,i}$$
$$= \prod_{i=1}^{n} \sum_{l=0}^{\infty} \frac{1}{l!} \left(-\frac{\beta\Gamma}{m}\right)^l \tilde{\sigma}'_{j,i}^T \{(\mathbb{E}_K - 1_K)\}^l \tilde{\sigma}_{j+1,i}. \quad (28)$$

$\tilde{\sigma}'^T_{j,i} \{\{(\mathbb{E}_K - 1_K)\}^l\} \tilde{\sigma}_{j+1,i}$ is calculated as

$$\tilde{\sigma}'^T_{j,i} \{\{(\mathbb{E}_K - 1_K)\}^l\} \tilde{\sigma}_{j+1,i}$$
$$= \tilde{\sigma}'^T_{j,i} \left\{\mathbb{E}_K + \frac{1}{k}\left\{(1-k)^l - 1\right\} 1_K\right\} \tilde{\sigma}_{j+1,i}$$
$$= \left\{\delta(\tilde{\sigma}'_{j,i}, \tilde{\sigma}_{j+1,i}) + \frac{1}{k}\left\{(1-k)^l - 1\right\}\right\}. \quad (29)$$

Thus, we have

$$\sigma_j'^T e^{-\frac{\beta}{m}\mathcal{H}_q} \sigma_{j+1}$$
$$= \prod_{i=1}^{n} \sum_{l=0}^{\infty} \frac{1}{l!} \left(-\frac{\beta\Gamma}{m}\right)^l \tilde{\sigma}'^T_{j,i} \{(\mathbb{E}_K - 1_K)\}^l \tilde{\sigma}_{j+1,i}$$
$$= \prod_{i=1}^{n} \left\{e^{-\frac{\beta\Gamma}{m}} \delta(\tilde{\sigma}'_{j,i}, \tilde{\sigma}_{j+1,i}) + \frac{1}{k} e^{-\frac{\beta\Gamma}{m}(1-k)} - \frac{1}{k} e^{-\frac{\beta\Gamma}{m}}\right\}$$
$$= \prod_{i=1}^{n} \left\{a\delta(\tilde{\sigma}'_{j,i}, \tilde{\sigma}_{j+1,i}) + b\right\} = b^n e^{s(\sigma'_j, \sigma_{j+1}) \log(\frac{a+b}{b})}.$$
$$(30)$$